# Explainable Text-Driven Neural Network for Stock Prediction

**Linyi Yang[1], Zheng Zhang[2], Su Xiong[2], Lirui Wei[2], James Ng[1], Lina Xu[2,3], Ruihai Dong[1,2]**

[1]Insight Centre for Data Analytics, University College Dublin, Dublin, Ireland
[2]Beijing Dublin International College, University College Dublin, Dublin, Ireland
[3]School of Computer Science, University College Dublin, Dublin, Ireland
{linyi.yang, james.ng, ruihai.dong}insight-centre.org
{zheng.zhang, su.xiong, lirui.wei}ucdconnect.ie
{lina.xu}ucd.ie

**Abstract:** It has been shown that financial news leads to the fluctuation of stock prices. However, previous work on news-driven financial market prediction focused only on predicting stock price movement without providing an explanation. In this paper, we propose a dual-layer attention-based neural network to address this issue. In the initial stage, we introduce a knowledge-based method to adaptively extract relevant financial news. Then, we use an input attention to pay more attention to the more influential news and concatenate the day embeddings with the output of the news representation. Finally, we use an output attention mechanism to allocate different weights to different days in terms of their contribution to stock price movement. Thorough empirical studies based upon historical prices of several individual stocks demonstrate the superiority of our proposed method in stock price prediction compared to state-of-the-art methods.

**Keywords:** Stock Prediction, Attention Mechanism, Explainable Model

## 1 Introduction

Stock market prediction has been an active area of strong appeal for both academic researchers and industry practitioners for a long time. The Efficient Market Hypothesis (Fama, 1965) states that stock market prices are largely driven by new information. Many recent works have shown success in predicting stock price movement based on text information (Ding et al., 2014; Ding et al., 2015; Li et al., 2013). Some events reported in news will influence people's decision-making which will essentially affect their trading behavior. It is believed that events reported in news have an impact on trading behavior which leads to fluctuations in stock markets. For example, Figure 1 shows the Dow Jones Industrial Average tumbled after US-China trade war fears appeared. Prices of individual stocks are influenced by relevant news. For instance, Steve Jobs once again sick leave resulting in Apple's stock price fell more than 4%. It is reasonable to state that information shapes stock movements.

Recently, a deep convolutional neural network for stock prediction has been proposed based on event embedding. (Ding, 2015) However, it has two main issues. First, it pays little attention to explaining the reason of the stock price prediction. Second, the events are extracted from news text using natural language processing toolkits, such as OpenIE (Fader, 2011) and dependency parsing. Obviously, the errors generated by natural language processing (NLP) tools will propagate in these methods. Furthermore, it will lose a large amount of useful information from those news which cannot be dealt with, since the limitation of existing NLP tools.

To solve the above problems, we propose a dual-level attention mechanism based on gates recurrent units network (GRU) to predict stock price movement, then give the explanations by looking back upon the most influential events over the last 7-days. In order to alleviate problems caused by noise in financial news datasets and obtain more discriminative news representations, we first build an input attention mechanism to allocate different weights to different news within one day in terms of their contributions to stock price movement. Then, we gather the news over the last seven days as input and pass to a GRU network. Third, we allocate different weights for different hidden units which indicates the output of a day. In this way, we can use the output of attention weights associated with news input to explore what leads to the stock price fluctuation.

The existing large-scale knowledge bases such as Freebase (Bollacker et al., 2008), DBpedia (Auer et al., 2007) and YAGO (Suchanek et al., 2007) have been built and widely used in many NLP tasks, including web search and relation extraction. But few researches have leveraged knowledge graphs to extract relevant news information for stock price prediction. However, the majority of available news is irrelevant when one is only interested in predicting the price movement of a single stock. On the other hand, news articles which do not explicitly mention target company may be valuable for predicting stock price movement. By leveraging the power of knowledge graphs, we are able to take advantages of news articles on Intel, Google, etc. when predicting the stock price of Apple.

**Figure 1** Example news for the U.S. Dow Jones Industrial Average index

The contributions of this work can be summarized as follows:

- Our model pays more attention to look back upon the input news for getting explanations.
- Compared with existing neural network model for stock prediction, our model can make full use of all related news of the target company.
- In the experiments, we show that our dual-level attention mechanism is not only beneficial to serval individual stocks in the task of stock prediction, but also able to give human-understandable explanation for the results.

## 2 Methodology

In this section, we first introduce the notation we use in this work and the problem we intend to study. Then, we present a novel neural network that incorporates dual-level attention mechanism into a GRU to fulfill this task.

### 2.1 Notation and Problem Statement

Given a set of financial news titles T = $\{t_1, t_2 \dots t_n\} \in \mathbb{R}^{n \times d}$, where d is the dimension of the sentence embedding of the title which appear in one day.

Stock price prediction is considered as a binary classification problem. Our model concatenates the news title from previous seven days $X_i = \{T_i^1, T_i^2 \dots T_i^7\} \in \mathbb{R}^{n \times d \times 7}$ to measure the probability of the target stock movement denoted. Also, the output of the attention layer is helpful to get a glimpse of how model works.

### 2.2 Overview

Inspired by the theories of human attention (Hübner et al., 2010) which argue that behavioral results are best modeled by a dual-stage attention mechanism. We present our attention-based neural network that incorporates a dual-stage attention mechanism. Figure 2 shows our neural network architecture which demonstrates the process that predicts the stock price movement for the next day. As shown in Figure2, our model contains six components:

1) Input layer: Original news titles within seven days before the prediction day.

2) Embedding layer: Each news title is mapped into a 300-dimension sentence embedding.

3) Input attention layer: The input attention enables the network to catch the important news and neglect the irrelevant news of one day by allocating different weights.

4) GRU layer: Features are extracted automatically for stock price prediction at this layer.

5) Output attention layer: Although the input attention can allocate different weights for different events at one day, we still need to take into account the long-term impact of the influential news which appears in previous days by output attention mechanism.

6) Output layer: Predict the stock price fall or raise for the next opening day.

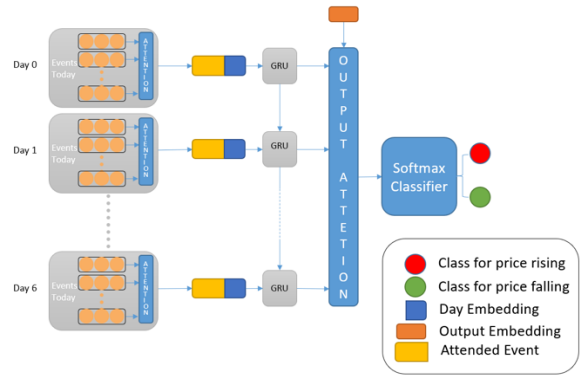

**Figure 2** The architecture of the GRU model with a dual-stage attention mechanism

### 2.3 Vector Representation

**Sentence Embedding.** Inspired by (Ding et al., 2015), using news titles as input performs better than using full news article in the task of text-driven stock price prediction. Therefore, in this work, the news titles are used directly. First, each title is tokenized and stop words are removed. After obtaining the tokens, each token is mapped to a pre-trained word embeddings by Spacy (Pennington et al., 2016). Finally, the average of all word embeddings is taken as the sentence embedding of the given news title. Sentence representations are encoded by column vectors in an embedding matrix T$\in \mathbb{R}^d$, where d is the dimension of the sentence vectors.

**Day Embedding.** We take advantage of time information by adding a day embedding to the input of the GRU. The day embedding is utilized to enhance the network to distinguish the different influence among days, which is denoted as D $\in \mathbb{R}^{d \times 7}$, where there are seven days to be embedded, and d is the dimension of the sentence vectors. D is a variable that can be trained jointly with the network. It is supposed to be trained into seven different variables that can represent the different influence of each day, which will be concatenated with the output of the input attention layer.

### 2.4 Input Attention Mechanism

Attention-based neural networks were first introduced by (Bahdanau et al., 2014) in machine translation. It has been applied in many NLP tasks in recent years, including relation extraction, recommend system, and knowledge completion. Dual-stage attention mechanisms have also been previously developed in time series prediction (Qin et al., 2017). In this section, we adopt an input attention for text-driven stock prediction tasks. Our attention mechanism aims to use input attention to allocate different weights to different news titles in terms of their contribution of influence the stock price movement.

First, the news titles from the previous seven days are passed through the input attention layer to obtain scores for each news title. The score of the news in day t is given by:

$$S_t = \text{softmax}(T_t W_s + b_s) \quad (1)$$

Where $T_t \in \mathbb{R}^{d \times n}$, $Ws \in \mathbb{R}^n$ which is the weight in this layer, and $b_s \in \mathbb{R}^d$.

Then, we compute the input to the GRU for day t, denoted as $x_t'$ by the weighted average of all news in one day.

$$x_t' = \frac{1}{n} \times \sum_{i=0}^{n-1} S_i T_i \quad (2)$$

Finally, the input of the cell in day t is given by:

$$x_t = (x_t', D_t) \quad (3)$$

where $x_t'$ is the weighted average of all news in one day and $D_t$ is the day embedding described above.

### 2.5 GRU Layer

Gated recurrent neural network is a simplified version of the long short-term memory neural network, but it has been proved performing better in many NLP tasks (Lin et al., 2016; Zhou et al., 2016). For the GRU layer, the output of the update gate is noted as U. The output of the reset gate is noted as R and the output of the cell is noted as h.

The output of the gate at day t is given by:

$$U_t = \sigma(W_u[h_{t-1}, x_t]) \quad (4)$$

$$R_t = \sigma(W_r[h_{t-1}, x_t]) \quad (5)$$

where σ represents the sigmoid function, $x_t$ and $h_{t-1}$ are the input vector and previous hidden state respectively, and Wr and Ur are weight matrices. The short-term memory of the cell is denoted as $h_t'$ and can be computed by:

$$h_t' = \tanh(W_h[R_t \circ h_{t-1}, x_t]) \quad (6)$$

The reset gate output $R_t$ position-wise multiplicity the output from the last cell $h_{t-1}$ and then cascade with the input $x_t$. For the output of this cell, known as the long-term memory of GRU network $h_t$, it is given by:

$$h_t = (1 - U_t) \circ h_{t-1} + U_t \circ h_t' \quad (7)$$

After all days are processed, the output from the last cell of GRU is regarded as the features for the previous seven days.

### 2.6 Output Attention Mechanism

The output embedding weight Wo $\in \mathbb{R}^d$ is a pattern that can be learned and the similarity weight $Ws \in \mathbb{R}^{d \times d}$ is a diagonal matrix comparing the similarity between the output for today and the output embedding pattern and gives the score for each day. The output day attention layer is shown in Figure 3.

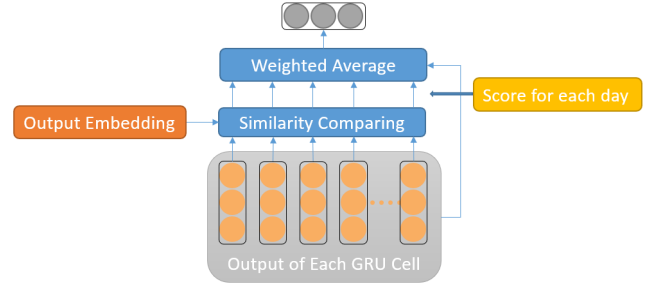

**Figure 3** The architecture of the output attention mechanism

The score $S_o$ of day t is obtained with the following equation:

$$S_o = h_t W_s W_o \quad (8)$$

Then, the output a can be computed by the weighted average of the output in seven days with the equation:

$$A = \frac{1}{7} \times \sum_{i=0}^{6} S_i h_i \quad (9)$$

### 2.7 Output Layer

The output layer determines whether the stock price will rise or fall. In practice, we calculate the conditional probability through a softmax function as:

$$y = softmax(W_{cls} A + b_{cls}) \quad (10)$$

Where $W_{cls}$ is the representation matrix of stock price movement, and $b_{cls} \in \mathbb{R}^2$ is a bias vector.

Inspired by (Ding et al., 2015), we employ a loss function using cross-entropy. We adopt the Adaptive Moment Estimation (Adam) (Kingma et al., 2015) update rule to learn parameters by minimizing the loss function.

Furthermore, in order to prevent overfitting, we apply dropout (Srivastava et al., 2014) on the output layer which aims to achieve better performance by randomly dropping out neural units during the training phase. Then, the output of our model is computed as follows:

$$y = softmax(W_{cls} A \circ h + b_{cls}) \quad (11)$$

where the vector h contains Bernoulli random variables with probability p.

## 3 Experiment

Our experiments aim to illustrate that our deep neural networks with dual-stage attention can not only predict the stock price movement, but also give explanation for the prediction. In this section, we first specify our settings and describe our datasets. Next, we compare the performance of our model on a widely used dataset with existing neural network architectures. Finally, we show that our approach, GRU+2ATT, can effectively outperform the most neural network models. Meanwhile, the output of our customized attention mechanism is able to give explanation for the prediction.

### 3.1 Data

We use the news of Reuters and Bloomberg in financial sector from 10/10/2006 to 26/11/2013 which is developed by (Ding et al., 2014) to predict the stock price of several target companies. We select three different companies from S&P500, including Apple, Google, and Minnesota Mining and Manufacturing (3M). End of day stock prices of these companies are collected.

For each company, we first select the news which contain this company's name directly, then we extend the news dataset by selecting relevant news. Specifically, we find the related entities automatically for each company which are stored in a known knowledge base, Wikipedia. Then the relative entities are used as key words to search the relevant news. We illustrate our dataset in detail in the following Table 1. Also, the divide of the dataset is shown in Table 2.

**Table I** The financial news dataset used for experiments

| Companies | | | |
|---|---|---|---|
| Company | 3M | Google | Apple |
| News | 504,448 | 130,195 | 53,782 |
| Period | 10/10/2006 – 14/11/2013 | | |

**Table II** Split the dataset

| Partition Strategy | | | |
|---|---|---|---|
| Category | Training Set | Testing Set | Validation Set |
| Days | 1480 | 180 | 180 |
| Period | 10/10/2006 – 18/06/2012 | 19/06/2012 – 07/03/2013 | 08/03/2013 – 14/11/2013 |

### 3.2 Evaluation Metrics

As in (Ding et al., 2015), we apply Matthews Correlation Coefficient (MCC) to evaluate our model in stock prediction. Here we report the precision and the MCC score. MCC is calculated as:

$$\frac{TP \times TN - FN \times FP}{\sqrt{((TP + FP)(TP + FN)(TN + FP)(TN + FN))}} \quad (12)$$

which indicates the quality of the binary classification in case there is bias between two classes.

### 3.3 Experiment Setup

**Sentence Embedding.** We map our news titles into a sentence embedding by Spacy, which provides an open-source develop platform. We fix the sentence length in 100 words, and then generate the sentence embedding in 300 dimensions.

**Parameter Settings.** We illustrate our hyper-parameters used in out experiment in Table 3.

**Table III** Hyper parameter settings

| Settings | |
|---|---|
| Sentence embedding dimension | 300 |
| Day embedding dimension | 5 |
| Batch size | 20 |
| Dropout probability | 0.5 |

### 3.4 Performance

To evaluate the proposed method, we compare our approach against two methods which were developed by (Ding et al., 2015). We reproduce their model by constructing the models revealed on their paper without the event extraction.

**Baselines:**

- EB-NN: sentence embeddings input and standard neural network prediction model
- EB-CNN: sentence embeddings input and convolutional neural network prediction model

**Experimental results compared with baselines:**

The result of the two baseline models and the proposed models are compared in Table 4.

**Table IV** Experimental Results

| Company | 3M ACC | MCC | Google ACC | MCC | APPLE ACC | MCC |
|---|---|---|---|---|---|---|
| EB-NN | 0.62 | 0.24 | 0.58 | 0.12 | 0.54 | -0.02 |
| EB-CNN | 0.68 | 0.30 | 0.60 | 0.14 | 0.58 | 0.11 |
| GRU-2ATT | **0.74** | **0.47** | **0.68** | **0.47** | **0.62** | **0.21** |

From Table 4, we can see that our model outperforms two baselines which are without attention mechanism for three companies from different areas in terms of both accuracy and MCC score. It indicates that the proposed GRU-2ATT is beneficial. The reason is that the input attention would dynamically focus on the more informative news while the output attention would allocate different weights for sentence embedding in different days.

### 3.5 Case Study

As compared with the previous work (Ding et al., 2015), we extend our dataset for individual company by searching in large-scale knowledge bases, which brings us more relative information for stock prediction.

When analyzing the risk of rising or falling stocks of listed company Google, we can find that YouTube is a subsidiary company of Google through the financial knowledge graph we have constructed which is shown in Figure 4. When major news of YouTube or YouTube's important affiliates has a major news, it will affect Google's share price.

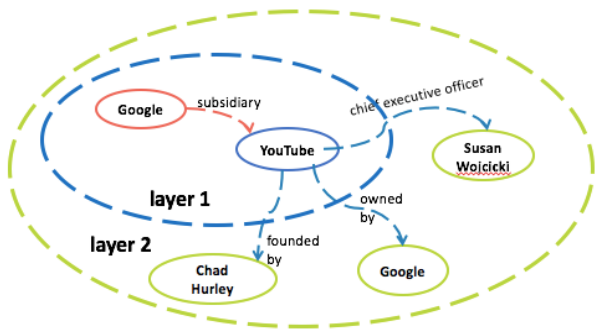

**Figure 4** The relative entities for the Google

Table 5 shows the output of our output attention mechanism for predicting the Google shares on May 4, 2007. Our prediction output benefits from expanding our knowledge of Google's news through knowledge graph shown in Figure 4.

**Table Ⅴ** The output of the customized attention mechanism

| | |
|---|---|
| Yahoo declines comment on reports of Microsoft talks. | 0.012 |
| Yahoo shares rise on reports of Microsoft interest. | 0.024 |
| **Premier League soccer sues YouTube over copyright.** | **0.514** |

From Table 5, we can note that the news "British Football Super League because of copyright issues to sue YouTube." has the strongest influence Google's share price. Furthermore, our model has been proved that it can benefit from the extended dataset. Since the 'YouTube', 'Microsoft', and 'Yahoo' are relative companies we map from the existing knowledge base, Wikipedia.

On May 4, 2007, we just select three news to give an instance of the explainable model. As a result, on the next opening day, Google's stock price suffered a big drop due to the prosecution. The experimental result has shown that our model can not only achieve the state-of-the-art performance on a widely used dataset, but also can output a reasonable explanation benefit from a dual-stage attention mechanism.

## 4 Conclusions

We have presented a novel dual-stage attention mechanism with a GRU. The method makes full use of the original news dataset, we use an input attention mechanism to reduce the noisy news for each day, and also adopt an output attention mechanism to allocate different weights for the seven days before the prediction day. Also, our method has a much better explainable result than the traditional methods. Experimental results show that our model outperforms than the baselines and also can give a reasonable explanation.

## Acknowledgements

This research was supported by Science Foundation Ireland (SFI) under Grant Number SFI/12/RC/2289.